\documentclass[letterpaper, 10 pt, conference]{ieeeconf}  %

\IEEEoverridecommandlockouts                              %

\usepackage{amsmath} %
\usepackage{amssymb}  %

\usepackage{bm}  %
\usepackage{graphicx}  %
\usepackage{subcaption} %
\usepackage{siunitx}
\usepackage{algpseudocode} %
\usepackage{algorithm} %

\usepackage{flushend} %
\usepackage[T1]{fontenc} %
\usepackage[utf8]{inputenc}
\usepackage{wrapfig} %

\usepackage[
        backend=biber,
        style=numeric,
        sorting=none,
        doi=false,
        isbn=false,
        url=false]{biblatex}

\newcommand{\dbm}[1]{\dot{\bm{#1}}}

\newcommand{\eq}[1]{Eq.~(\ref{#1})}
\newcommand{\fig}[1]{Fig.~\ref{#1}}

\title{\LARGE \bf
        Learning Human-Inspired Force Strategies for Robotic Assembly
        }

\author{Stefan Scherzinger$^{1}$, Arne Roennau$^{1}$ and R\"udiger Dillmann$^{1}$%
\thanks{
 ${}^{1}$ All authors are with FZI Research Center for Information Technology, Haid-und-Neu-Str. 10-14, 76131 Karlsruhe, Germany
        {\tt\small \{scherzinger, roennau, dillmann\}@fzi.de}}%
}

\begin{document}

\maketitle
\thispagestyle{empty}
\pagestyle{empty}

\begin{abstract}
The programming of robotic assembly tasks is a key component in manufacturing and automation.
Force-sensitive assembly, however, often requires reactive strategies to handle
slight changes in positioning and unforeseen part jamming.
Learning such strategies from human performance is a promising approach, but
faces two common challenges:
the handling of low part clearances which is difficult to capture from demonstrations
and learning intuitive strategies offline without access to the real hardware.
We address these two challenges by learning probabilistic force
strategies from data that are easily acquired
offline in a robot-less simulation from human demonstrations with a
joystick.
We combine a Long Short-Term Memory (LSTM) and
a Mixture Density Network (MDN) to model human-inspired behavior in such a way
that the learned strategies transfer easily onto real hardware.
The experiments show a UR10e robot that completes a plastic assembly with
clearances of less than 100 micrometers whose strategies were solely demonstrated in simulation.

\end{abstract}

\section{Introduction}  %

Many industrial assembly processes require mounting together individual components to build complex products.
When plug-inserting two individual parts, this is referred to as the
peg-in-hole problem and is often used to benchmark new methods in robotics research.
In industry, offline programming~\cite{Pan2012} is an established process of
deriving robot programs before fine-tuning and deploying them on a plant.
This process of \textit{fine motion planning}~\cite{Gottschlich1994} is
particularly challenging for force-sensitive assembly:
Stiff contacts and rigid fixtures can build up high reaction forces between
the robot and the environment, and usually require careful calibration of the robots'
motion to process parameters such as friction.
On-site, expert knowledge is often needed due to the complexity of the programming.

In general, the inclusion of force-torque sensor signals seems crucial for
assembly success \cite{Newman1999}, \cite{InWook1999}, \cite{Newman2001} and
is still used today in form of analytical solutions \cite{Tang2016}, \cite{Wahrburg2015}.
Advances in lightweight robots %
enable a task-optimized impedance design and achieve robust assembly skills that are comparable to human performance
\cite{Stemmer2007}, \cite{Johannsmeier2019}.
Using the force-torque signals for control policies can also be learned by the robots from
trial-and-error~\cite{Levine2015}, which has been shown for
insertion tasks with tight clearances \cite{Inoue2017}, \cite{Schoettler2019}.
On the contrary, humans are particularly good at handling these tasks manually
by using sophisticated senses with a high bandwidth of information.
Although being intuitive for us where to press, turn and jolt to get two parts mated when using our hands,
these strategies are fuzzy and difficult to describe quantitatively for robot programs.
Capturing and translating these skills into suitable algorithms is an exciting challenge.
In this paper, we consider offline programming for assembly, where motion is highly constrained
and the parts have low clearances and tight fits.
The principle idea is to find technical representations for these human-inspired skills
that can be transferred to real hardware once programmed.

To work toward this goal, we propose \textit{probabilistic force strategies},
a data-driven approach to mimic human behavior for low-clearance
assembly skills that we build onto force control on industrial manipulators.
Our proposed deep neural networks learn from data that is obtained easily offline
through showing the task in simulation on simplified primitives with a joystick.
Building on the experience from our earlier approach \cite{Scherzinger2019Contact},
we redesign the previous neural network architecture and present
a new concept for modeling the quintessence of human behavior especially for low-clearance assembly.
\section{Problem statement and Related Work}
\label{sec:related_work}

The peg-in-hole problem is well-known in the assembly literature.
Shapes and sizes vary but the parts usually share a single insertion direction.
In addition to low tolerances and tight fits,
we extend this to various insertion directions with higher geometric complexity in contact.
Low-clearance assembly means dealing with force closure and form closure
adaptively during execution, caused by inevitable model inaccuracies.
The additional challenge in offline programming, however, is programming these skills without real hardware.

We follow the core idea to learn these skills from human demonstrations, %
which is known under Programming by Demonstration (PbD) \cite{Dillmann1995}, \cite{Billard2008},
Imitation Learning (IL) \cite{Schaal1999} in physiological neurosciences, and Behavioral Cloning (BC)
\cite{Bain1995} from machine learning in general.
The following sections discuss related works.
\subsection{Probabilistic skill learning from demonstration}
Gaussian Mixture Regression (GMR) is one of the frequently used methods for trajectory learning in
motion and force-based manipulation \cite{Hersch2008},
\cite{Calinon2009}, \cite{Rozo2016}.
Tang et al.~\cite{Tang2016} use this approach to describe a probabilistic mapping from
contact wrenches to twist commands for admittance control.
Although using probabilistic modeling for approximating the datasets, the
regression from these distributions is usually deterministic.
Kronander et al.~\cite{Kronander2014} use sampling from the conditional
probability densities to introduce randomness in comparison to Least-Squares Estimates (LSE) and thus
overcome deadlocks during assembly.
Note that this is mostly valid only if the samples need not be coherent in time.

Al-Yacoub et al.~\cite{Al-Yacoub2019} use GMR
during peg-in-hole assembly tasks for clustering force-torque
signals and getting insights into human-applied contact transitions and skill characteristics.

Dong et al.~\cite{Dong2007} combine a Hidden Markov Model (HMM) with Locally Weighted Regression (LWR)
to learn discrete contact states of peg-in-hole assembly in a haptic simulation environment.
During reproduction, they use LWR for angle corrections based on forces-torque signals in these states.
Vergara et al.~\cite{Vergara2019} use
Probabilistic Principal Component Analysis (PPCA) %
for modeling kinesthetic teachings.
Combining this with the constraint specifications of an optimization framework,
they obtain partially transferable skills to new target positions, in which
imitation learning covers wider workspace motions and constraint-based optimal
control considers the last centimeters.

\subsection{DMPs with force profiles}

Dynamic Movement Primitives (DMP)~\cite{Ijspeert2013}
are an established method for encoding human-recorded motion profiles into parametric models.
The inclusion of force profiles makes them suitable for contact-rich tasks, such as assembly~\cite{Abu2015}.
Methods for teaching include e.g. kinesthetic guiding
\cite{Kramberger2016} and teleoperation \cite{Krueger2014},
\cite{Savarimuthu2017}.
The latter has advantages for obtaining unbiased force profiles for not having to move the robot in its morphology.
The Force profiles are often not embedded into the transformation system of the DMP, but instead are parameterized and used directly as reference setpoints in end-effector force controllers.
They can be tracked in parallel to the motion trajectories, e.g. with a proportional-integral (PI) controller \cite{Abu2015}.
A compliance matrix can map the force-torque reference setpoints to motion setpoints for velocity-controlled robots.
Nemec et al.~\cite{Nemec2013} and succeeding works \cite{Abu2015},
\cite{Krueger2014}, \cite{Kramberger2017}, \cite{Savarimuthu2017}
apply this mechanism to peg-in-hole assembly tasks.
Abu-Dakka et al.~\cite{Abu2015} and Savarimuthu et al.~\cite{Savarimuthu2017}
further use the DMP's phase modulation to slow down the execution speed when
losing track of the target force profiles in contact.
\subsection{Our approach}
Tackling force-based assembly from within offline programming requires simulation environments.
Compared to using real hardware, simulation offers the advantage of supporting
demonstrations with ease but usually poses high demands on realism to successfully transfer the skills to real hardware.
Onda et al.~\cite{Onda1995}, used a robotic manipulator as the teach device to steer
objects in a frictionless simulation.
Their concept did not include force feedback and purely relied on the visual channel
but was sufficient to extract contact states for deriving simple position/force commands.

Applying more recent methods to this approach, we propose \textit{sequential-probabilistic}
models for encoding these complex, human strategies for
low-clearance assembly, using 
the combination of a Mixture Density Network (MDN)~\cite{Bishop1994} with
Long Short-Term Memory (LSTM)~\cite{Hochreiter1997}.
For execution on real hardware, we connect this new skill model to an admittance
control scheme for industrial manipulators.

\section{Human-Inspired Force Strategies}  %
\label{sec:contact_handling}

Klingbeil et al. investigated human behavior in assembly tasks~\cite{Klingbeil2017} and
tested the hypothesis that humans ground assembly strategies in mental representations of contact states.
Their results showed that some users passed a few common states during
teleoperation but the intermediate behavior was mostly random among the participants.
Physiological studies on control and planning through the central nervous system
further showed that humans
use mental \textit{forward models}~\cite{Jordan1992}
to predict the response to their motor commands~\cite{Miall1996}.

Concluding from these results,
we propose to drop the high-level construct of contact states and model human-inspired assembly skills
directly from the cyclic interaction between teleoperated
force-torque commands and observed relative motion.

\subsection{Learning features and simulation}

\begin{figure*}
        \centering
        \includegraphics[width=0.95\textwidth]{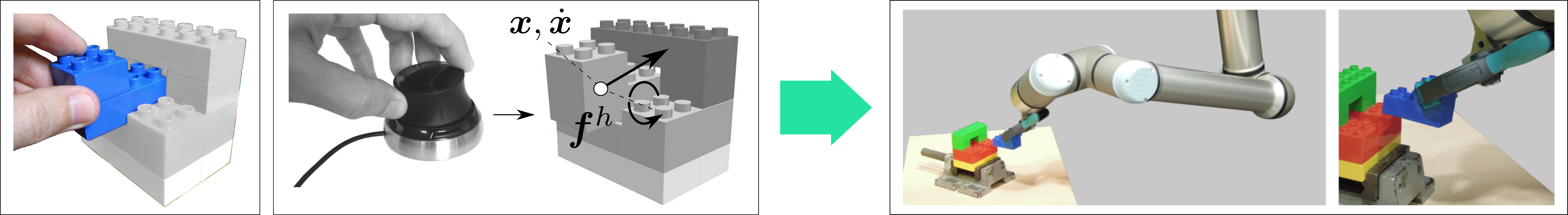}
        \caption{
                From simulated teleoperation to robot-controlled execution.
                We use a 3Dconnexion\texttrademark  ~3D joystick to steer the
                assembly objects and solve difficult jamming situations.
                An LSTM-MDN model learns reactive strategies from recorded
                demonstrations and controls the robotic manipulator during the real assembly.
        }
        \label{fig:mental_forward_models}
\end{figure*}
We apply this insight to human-inspired assembly skills as follows:
A user teleoperates an assembly in simulation with a joystick.
In our example, the assembly consists of two plastic components, that are to be joined together with dexterity.
A low clearance between the parts and complicated geometry for insertion make
this a challenging benchmark, which we use to describe the concepts of this
paper.
\fig{fig:mental_forward_models} illustrates the overall concept.
In the simulator, we record the human-initiated vector $\bm{f}^h \in
\mathbb{R}^{6}$ that we apply with our fingers via the joystick to steer one of
the objects.
To obtain this signal, we interpret the joystick's six-axes displacement as a
combined vector of forces and torques and scale that to suitable ranges for the
real application.
We build the simulation environment on our previous
work~\cite{Scherzinger2019Contact}, in which we describe further requirements,
such as velocity-dependent damping while steering the assembly objects.
Increased friction and stiction between the assembly components during demonstrations shall
lead to robust strategies for the transfer to the real setup.
We also record the combined position and orientation vector $\bm{x} \in
\mathbb{R}^{7}$, and the goal-relative velocity $\dbm{x} \in \mathbb{R}^{6}$
during assembly.
In contrast to our previous simulator, we describe both variables with respect to the estimated goal pose.

We believe that the interaction between 
the recorded motor commands $\bm{f}^h$
and the corresponding outcome $\bm{x}, \bm{\dot{x}}$ holds concise information for learning human-inspired strategies,
and the goal of this paper is to present an improved neural network controller that mimics
these strategies for low-clearance assembly.

To keep equations short, we introduce two abbreviations:
\begin{align}
        \label{eq:label_abbreviations}
        \hat{\bm{x}}_t &=  \left[\bm{x}_t \dbm{x}_t {\bm{f}_t^h} \right]^T \\
        \hat{\bm{y}}_t &= \bm{f}_{t+1}^h~,
\end{align}
in which $\hat{\bm{x}}_t$ is the concatenated state of all variables at the time
step $t$ and $\hat{\bm{y}}_t$ is the force-torque vector at the succeeding time
step.
\subsection{Data analysis}

\begin{figure*}
        \centering
        \begin{subfigure}[b]{0.49\textwidth}
                \includegraphics[width=1.0\textwidth]{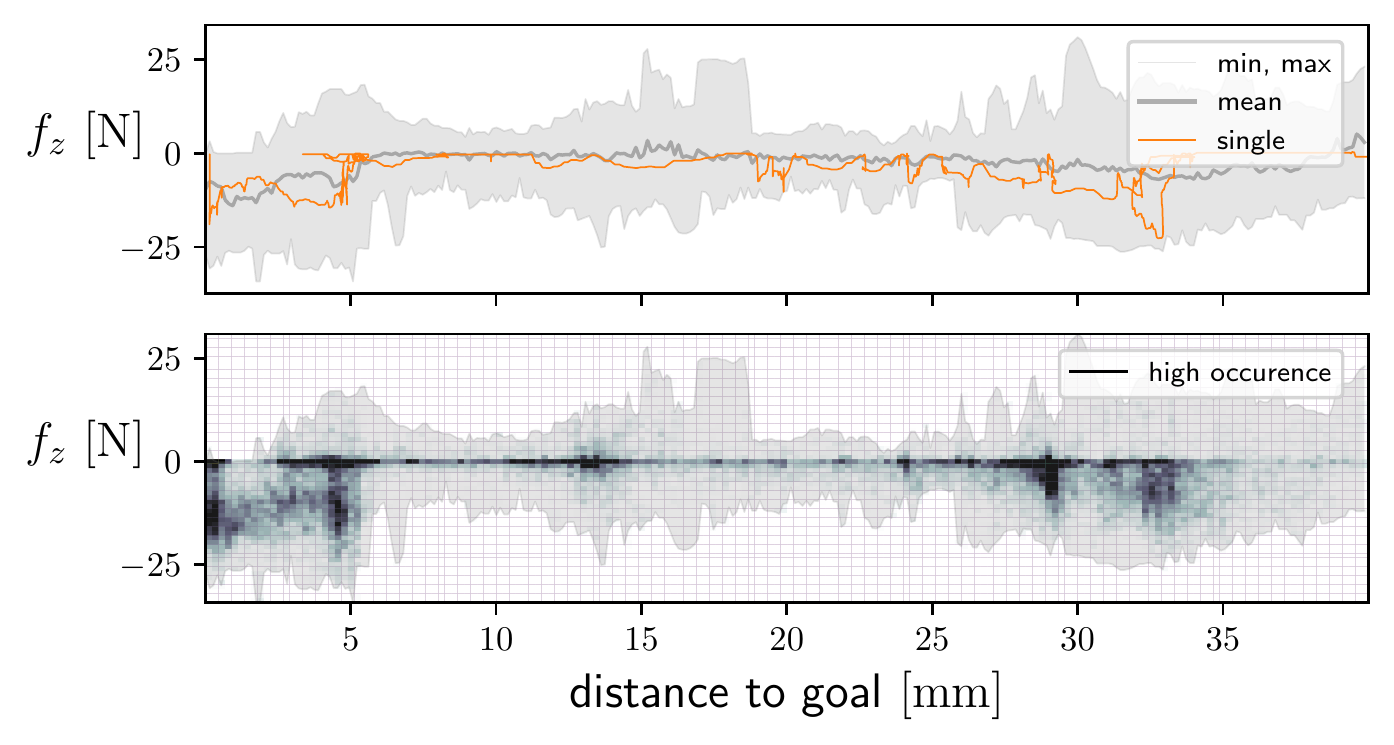}
                \caption{}
        \end{subfigure}%
        \begin{subfigure}[b]{0.49\textwidth}
                \includegraphics[width=1.0\textwidth]{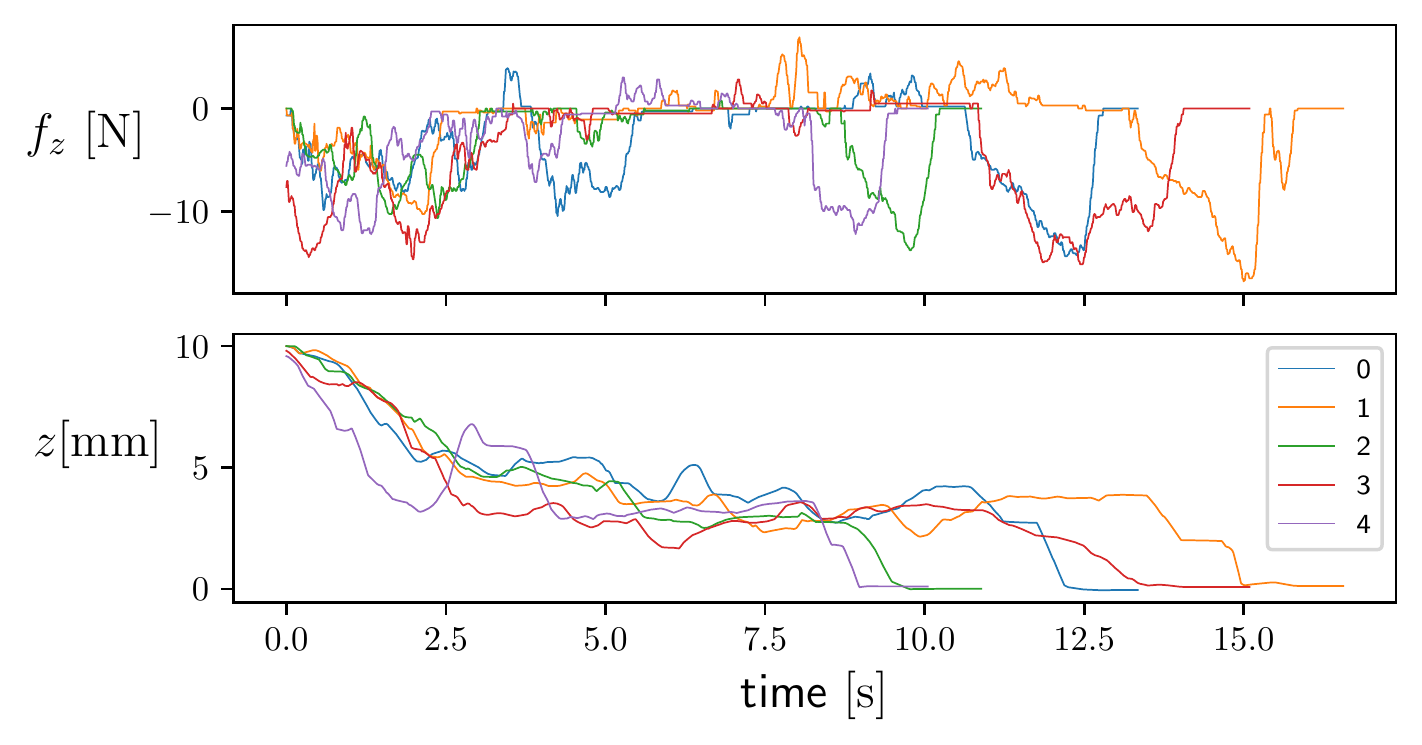}
                \caption{}
        \end{subfigure}
        \caption{
                Analysis of recorded demonstrations in the simulator.
                (a)~\textit{above} Band plot of the $f_z$ component during 100 demonstrations.
                We used statistical data binning to cluster all demonstrations into vertical bins,
                from which we computed the min, max, and mean values, respectively.
                An additional, single demonstration visualizes strategies during assembly.
                (a)~\textit{below} The color gradient shows the qualitative
                occurrence of discretized values for forces $f_z$ in a $180 \times 60$ grid.
                (b)~\textit{above} Time sequences for 5 performances reveal repeating patterns in $f_z$.
                (b)~\textit{below} The vertical position $z$ shows the assembly’s progress toward zero, which
                indicates the end of the assembly. 
        }
        \label{fig:training_data}
\end{figure*}

Analyzing the recorded data gives us insights into how to design the required learning algorithm.
\fig{fig:training_data}(a)~\textit{above} shows the z-component $f_z$ of the motor commands $\bm{f}^h$ for the exemplary assembly.
This component is the vertical pushing force and is one of the characteristic
variables when joining both parts at the end of the assembly.
We use the Euclidean goal distance as an indicator of assembly progress.
For tasks with low clearances, this distance is mostly unambiguous in geometry and provides
a one-dimensional reference.
The plot shows that it will not be possible to ground the force-based strategies solely in position for this assembly,
because the mean loses the characteristics of the individual executions.
\fig{fig:training_data}(a)~\textit{below} shows the respective histogram of the z-component $f_z$.
The darker the spots, the more frequently do these values appear in the recorded data.
This also indicates in which regions it took longer to complete the assembly.
Having vertical, dark areas reveals an inverse learning problem:
For instance, at around \SI{5}{mm} goal distance, any $f_z$ between \SI{0}{N} and \SI{-25}{N} seems valid, because
these data appear frequently in this region in the data set.
In conclusion, we need a probabilistic model to capture the non-deterministic components of $f_z$.

Fig.~\ref{fig:training_data}(b)~\textit{above} shows $f_z$ over time for five individual recordings.
The curves share patterns in specific time windows that slightly vary in scale and offset.
For instance, there are characteristic pushing profiles at the end of the assembly that
start at around \SI{8}{s} for the fastest demonstration and at around \SI{13}{s} for the
slowest demonstration.
While a probabilistic model could reproduce the distributions from
Fig.~\ref{fig:training_data}(a)~\textit{below}, it could not generate predictions coherent in time as they appear
in our recordings.
In conclusion, we additionally need a sort of memory to model sequential patterns in the human-recorded strategies.

\section{Force Strategies - a Temporal-probabilistic Approach}
\label{sec:temporal_probabilistic_approach}

The previous section motivated two key components in modeling force strategies for our benchmark:
probabilistics to cover the multivalued character of the distributions, and
memory to model their time dependencies.
In this section, we describe how we include both
into a temporal probabilistic model for learning force strategies for low-clearance assembly.

\subsection{Capturing patterns in sequences}
We build our model on the Long Short-Term Memory (LSTM)~\cite{Gers2000}, which has
been used in many domains and which excels at learning long-term dependencies in data.
In contrast to feed forward networks, LSTMs maintain an internal state,
the cell state $c_t$, and update information via gated operations.
This allows them to effectively encode long sequences of data.
In our work, we use the LSTM primarily to encode time windows of past assembly progress into a concise
representation and will later combine this with a probabilistic output layer to
predict meaningful next force strategies with high confidence.

We combine $m$ of these cells into an LSTM layer, allowing us to describe the forward
pass with the following matrix-based equations~\cite{Hochreiter1997}, \cite{Gers2000}:

\begin{equation}
        \label{eq:lstm_equations}
        \begin{aligned}
                \bm{f}_t &= \text{sigmoid} ( \bm{W}_f \hat{\bm{x}}_t + \bm{b}_f + \bm{U}_f \bm{h}_{t-1}) \\
                \bm{i}_t &= \text{sigmoid} ( \bm{W}_i \hat{\bm{x}}_t + \bm{b}_i + \bm{U}_i \bm{h}_{t-1}) \\
                \bm{o}_t &= \text{sigmoid} ( \bm{W}_o \hat{\bm{x}}_t + \bm{b}_o + \bm{U}_o \bm{h}_{t-1}) \\
                \bm{c}_t &= \bm{f}_t \circ \bm{c}_{t-1} + \bm{i}_t \circ \text{tanh} ( \bm{W}_c \hat{\bm{x}}_t + \bm{b}_c + \bm{U}_c \bm{h}_{t-1}) \\
                \bm{h}_t &= \bm{o}_t \circ \text{tanh} ( \bm{c}_t) .
        \end{aligned}
\end{equation}

The vector $\hat{\bm{x}}_t \in \mathbb{R}^c$ is the input to the LSTM layer and $\bm{h}_t \in \mathbb{R}^m$ is the output. 
$\bm{f}_t, \bm{i}_t, \bm{o}_t \in \mathbb{R}^m$ are the activation vectors for the forget
gates, the input gates, and the output gates respectively.
Each has a non-recurrent weight matrix $\bm{W} \in \mathbb{R}^{m \times c}$ and a
recurrent weight matrix $\bm{U} \in \mathbb{R}^{m \times m}$ for the
previous hidden state $\bm{h}_{t-1} \in \mathbb{R}^m$.
We further use $\circ$ to denote the element-wise product.
The layer-wide input $\bm{h}_{t-1}$ generates recurrent connections between each cell in an LSTM layer.
We use
\begin{equation}
        \label{eq:sigmoid}
        \text{sigmoid}(x) = \frac{1}{1 + e^{-x}}
\end{equation}
as activation for each element in a vector and
squash inputs and outputs with the hyperbolic tangent.
The weight matrices $\bm{W}, \bm{b}$, and $\bm{U}$ include all learnable parameters.
We use a single-layered LSTM and adjust the number of learnable parameters
with the number of cells $m$ in that layer.

Note that \eq{eq:lstm_equations} yields $N+1$ consecutive hidden states and
cell states for an input sequence of consecutive points $\hat{\bm{x}}_{t-N},
\dots, \hat{\bm{x}}_t$.
We are only interested in the last pair $\bm{h}_t, \bm{c}_t$, (which we refer
to as the \textit{encoding}) for which the LSTM has seen the complete input
sequence:
\begin{equation}
        \label{eq:lstm_function}
        \text{LSTM}(\hat{\bm{x}}_{t-N}, \dots, \hat{\bm{x}}_t) := (\bm{h}_{t}, \bm{c}_t) ~.
\end{equation}
The next section shows how we use this encoding to estimate probabilities for the next force-torque vectors.

\subsection{Modeling the ambiguity in human behavior}
The recorded force strategies had seemingly random trial and error, resulting
in distributions for specific regions.
Sampling from these distributions alone will not be successful, since those
samples will not form coherent strategies over time.
Using the encoding from the LSTM, however, we can model more specific
distributions to sample from and thus consider past assembly progress.
We use the Mixture Density Network (MDN)~\cite{Bishop1994} to model these characteristics.
LSTMs and MDNs have been combined for modeling highly complex human performance, such as generating
human handwriting from digital text \cite{Graves2013} or creating full-body, freestyle dance motion \cite{Crnkovic2016}.

MDNs model conditional probability densities of the target data, conditioned on the input data~\cite{Bishop1994}.
In our notation, we describe this with $p(\hat{\bm{y}}_t \mid \bm{h}_t, \bm{c}_t)$, in which the
hidden state $\bm{h}_t$ and cell state $\bm{c}_t$ are the encoding of the input sequence according to~\eq{eq:lstm_function}.

For combining the LSTM encoding and the MDN output, we feed
the encoding into a fully connected layer with linear activation
\begin{equation}
        \label{eq:mdn_activation}
        \bm{z} = \bm{W}_z \left[\bm{h}^T, \bm{c}^T \right]^T .
\end{equation}
The weight matrix $\bm{W}_z \in \mathbb{R}^{ k(6+2) \times 2m}$ introduces additional learning parameters.
The next step is to split $\bm{z}$ into a tuple
$(\bm{z}^{\alpha} , \bm{z}^{\mu} , \bm{z}^{\sigma})$
and use the individual components to parameterize a Gaussian mixture model
of $k$ Gaussian distributions $\mathcal{N}_6(\bm{\mu}, \sigma^2)$ that are each $6$-dimensional.
The six dimensions derive from our intention to model the $6$-dimensional force-torque vector.

First, we compute normalized mixing coefficients for the Gaussians from $\bm{z}^{\alpha}$ with a softmax
according to
\begin{equation}
        \label{eq:mdn_alphas}
        \alpha_i = \frac{\text{exp}(z_i^{\alpha})}{\sum_{j=1}^k \text{exp}(z_j^{\alpha})} ~, \qquad i = 1 \dots k ~, \qquad \alpha_i \in \mathbb{R} ~.
\end{equation}
The centers of the Gaussians are computed from $\bm{z}^{\mu}$ with
\begin{equation}
        \label{eq:mdn_mus}
        \bm{\mu_{i}} = z_{ij}^{\mu} ~, \qquad i = 1 \dots k ~, j = 1 \dots 6 ~, \qquad \bm{\mu}_i \in \mathbb{R}^{6}
\end{equation}
and the scale parameters from $\bm{z}^{\sigma}$ with
\begin{equation}
        \label{eq:mdn_sigmas}
        \sigma_i = \text{exp}(z_i^{\sigma}), \qquad i = 1 \dots k ~, \qquad \sigma_i \in \mathbb{R} ~.
\end{equation}
Using these parameters, we can compute the conditional density for the $i$th Gaussian with~\cite{Bishop1994}
\begin{equation}
        \label{eq:conditional_density}
        \phi_i (\hat{\bm{y}}_t \mid \bm{\mu}_i, \sigma_i) =
        \frac{1}{(2\pi)^{6/2} \sigma_i^6} \text{exp}
        \left(- \frac{\lVert \hat{\bm{y}}_t - \bm{\mu}_i  \rVert ^2}{2 \sigma_i ^2} \right) .
\end{equation}
Finally, the conditional probability density is composed of a linear combination of $k$ Gaussian functions $\phi_i$~\cite{Bishop1994}
\begin{equation}
        \label{eq:probability_density}
        p(\hat{\bm{y}}_t \mid \bm{h}_t, \bm{c}_t) = \sum_{i=1}^{k} \alpha_i  \phi_i ( \hat{\bm{y}}_t \mid \bm{h}_t, \bm{c}_t) ~.
\end{equation}
This probability describes how likely $\hat{\bm{y}}_t$ follows the observed sequence $\hat{\bm{x}}_{t-n}, n = N \dots 0$,
which we feed in form of the LSTM's encoding $(\bm{h}_t, \bm{c}_t)$ into the MDN.
The computed mixture parameters thus become functions of the encoded sequence for each time step.
During training, this mechanism allows our model to solve ambiguity in the recordings by parameterizing
fine-grained probability density functions along the assembly progress.

\subsection{Data composition and training}
\label{sec:data_composition_and_training}
\begin{figure*}
        \centering
        \includegraphics[width=0.99\textwidth]{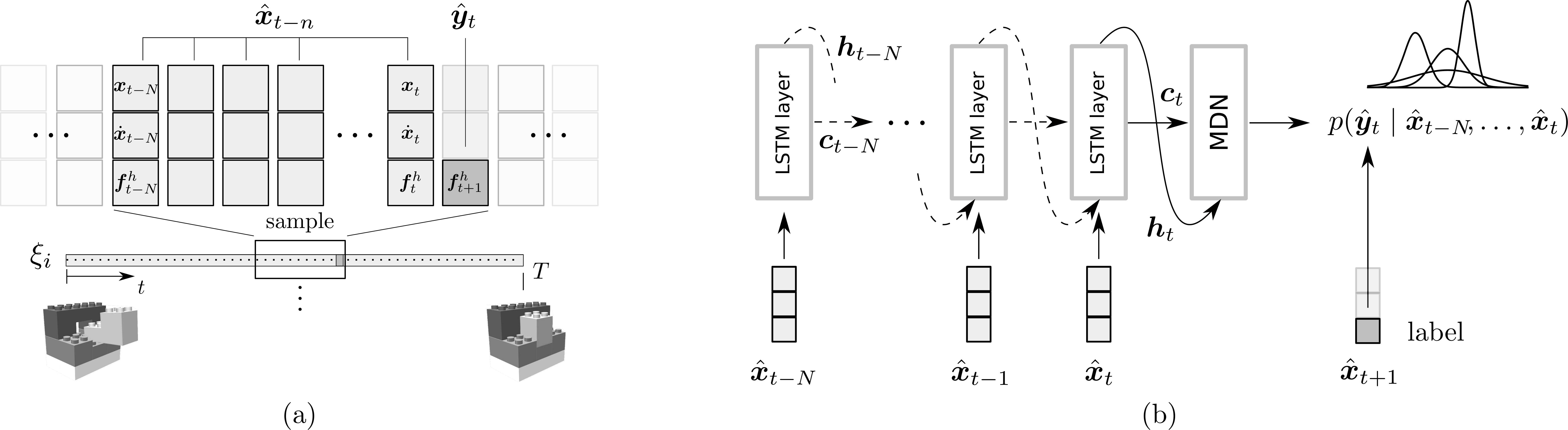}
        \caption{
        (a) Composition of training samples from recorded demonstrations in simulation.
        (b) The unrolled LSTM-MDN model with inputs and outputs during the
        forward pass. The model is trained with BPTT, using the negative
        log-likelihood of the conditional probability density.
        }
        \label{fig:data_composition_and_training}
\end{figure*}

The LSTM-MDN model is a recursive prediction model from state-action pairs to strategies.
Its desired output is motor control commands $\bm{f}^h$ that change the assembly object's state
$\bm{x}, \dbm{x}$, which in turn provokes a next motor command.
We use supervised learning for training by splitting the recorded
demonstrations in the simulator into input sequences and respective labels.
\fig{fig:data_composition_and_training}(a) illustrates this approach.
We obtain training samples from the data set by selecting individual demonstrations $\xi_i$ 
out of all recorded assembly demonstrations.
A pivotal time step $t$ selects the starting point.
We then compose the sample with the input sequence $\hat{\bm{x}}_{t-N}, \dots,
\hat{\bm{x}}_t$ and the respective label $\hat{\bm{y}}_t$, which is the
force-torque component of the next $\hat{\bm{x}}_{t+1}$.

The randomly extracted samples have an equal length of $N+2$, so that
the LSTM-MDN model has access to $N$ time steps before predicting the
corresponding probability density function.
During training, we sample batches for parallel processing and cover the
entirety of the training data stochastically.
We train our model with backpropagation through time (BPTT) by unrolling the model $N$ time steps.
\fig{fig:data_composition_and_training}(b) shows the unrolled model as a normal feedforward network.
The LSTM layer's recurrent connections become normal feedforward connections
and propagate the cell state and hidden state.
All LSTM instances from \fig{fig:data_composition_and_training}(b) share a
common set of learnable weights from \eq{eq:lstm_equations} and refer to the
same layer, which is illustrated at specific time steps together with the
respective input from the training sequence.
The fully connected output layer parameterizes the Gaussian mixture model, from which we
compute the conditional probability for the given label.
We finally optimize all learnable weights by minimizing the negative log-likelihood
\begin{equation}
        \label{eq:loss_function}
        \mathcal{L} = - \text{ln} \left( p (\hat{\bm{y}}_t \mid \hat{\bm{x}}_{t-N}, \dots, \hat{\bm{x}}_t) \right)
\end{equation}
of this conditional probability density.
\section{Force Control and Assembly Skills}
\label{sec:force_control}
\begin{figure}
        \centering
        \includegraphics[width=0.49\textwidth]{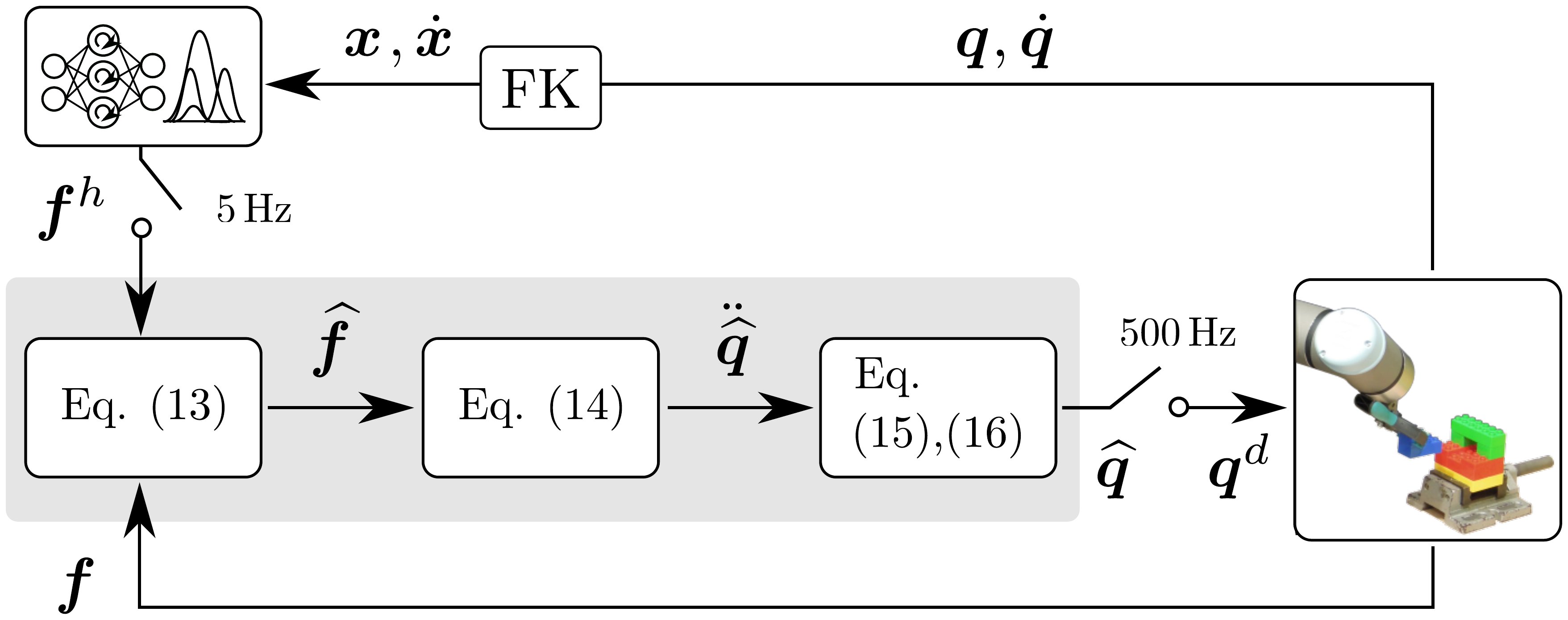}
        \caption{
        Control scheme of the neural network controller for assembly. The
        LSTM-MDN receives state feedback from the real robot and generates
        force-torque references for our simulation-based force control algorithm.
        }
        \label{fig:augmented_force_controller}
\end{figure}

During deployment, the LSTM-MDN model predicts 6-dimensional Gaussian
distributions, from which we sample force-torque strategies for robot control in each time step.
We use the forward dynamics-based control approach of our previous
work~\cite{Scherzinger2020virtual} to realize Cartesian force control on the robotic manipulator.
Its implementation is available
open-source\footnote{https://github.com/fzi-forschungszentrum-informatik/cartesian\_controllers}.
The algorithm applies the force-torque vector at the end-effector of a simulated robot
and computes the corresponding joint acceleration of the system with simplified forward dynamics.
The resulting motion is integrated and sent to the joint position servos of the real industrial robot.
\fig{fig:augmented_force_controller} shows the control scheme.

The hat symbol $\widehat{(.)}$ denotes \textit{simulated} values that
are computed on a simulated robot model with equal kinematics.
First, we compute an error vector from the reference force-torque profile $\bm{f}^h$ as
sampled from the LSTM-MDN model and the measurements $\bm{f}$ from a Cartesian
force-torque sensor in contact with the environment
\begin{equation}
        \widehat{\bm{f}} = \bm{K}_p ( \bm{f}^h - \bm{f} ) ~.
\end{equation}
$\bm{K}_p$ is a positive, diagonal gain matrix.
We then simulate the robot's response with
\begin{equation}
        \label{eq:forward_dynamics}
        \ddot{\widehat{\bm{q}}} = \widehat{\bm{H}}^{-1} \widehat{\bm{f}} ~,
\end{equation}
in which $\widehat{\bm{H}}$ is a linearized joint space inertia matrix that is computed in each control cycle.
We refer to our work \cite{Scherzinger2020virtual} for more details on this aspect.
We then obtain the joint positions $\widehat{\bm{q}}$ with the Euler forward method according to
\begin{align}
        \widehat{\bm{q}}_{t} = \widehat{\bm{q}}_{t-1} + \dot{\widehat{\bm{q}}}_{t-1} \Delta \widehat{t} \\
        \dot{\widehat{\bm{q}}}_{t} = \dot{\widehat{\bm{q}}}_{t-1} + \ddot{\widehat{\bm{q}}}_{t} \Delta \widehat{t} ~.
\end{align}
An additional joint damping $\dot{\widehat{\bm{q}}}_{t} \gets 0.9 ~
\dot{\widehat{\bm{q}}}_{t}$ in each simulation cycle avoids oscillations.
The simulated time window $\Delta \widehat{t}$ can be chosen independently of
the robot's real control rate and allows for tweaking the controller's response
to force-torque inputs.
The simulated joint positions $\widehat{\bm{q}}$ are then forwarded to the
joint position servos of the real manipulator.
Finally, we use a forward kinematics routine to compute the end-effector pose $\bm{x}$ and velocity $\dbm{x}$,
and feed them back into the LSTM-MDN model for the next prediction.
Note how model inference and force control run at different rates.
\section{Experiments and Results}
\label{sec:experiments}
\subsection{Dataset and simulation}
\begin{figure}
        \centering
        \begin{subfigure}[b]{0.49\textwidth}
                \includegraphics[width=1.0\textwidth]{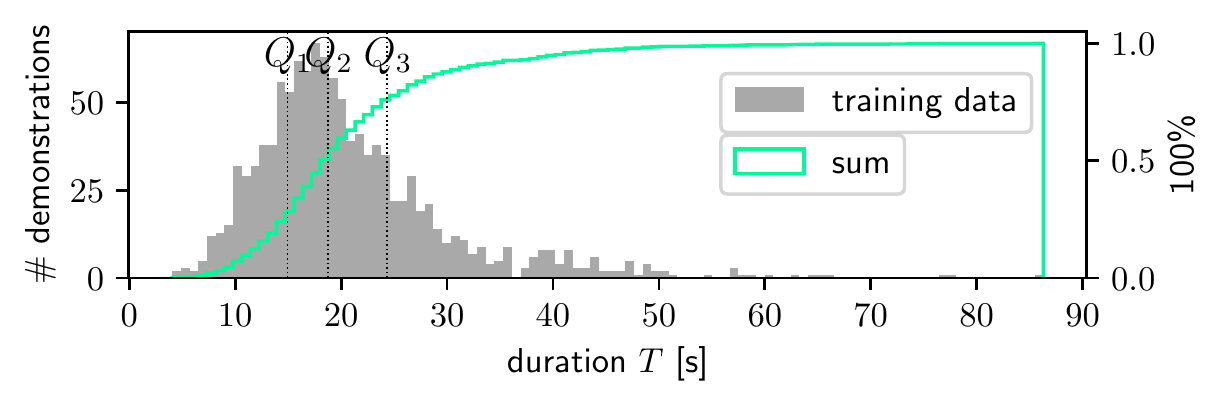}
                \caption{}
        \end{subfigure}
        \begin{subfigure}[b]{0.49\textwidth}
                \includegraphics[width=1.0\textwidth]{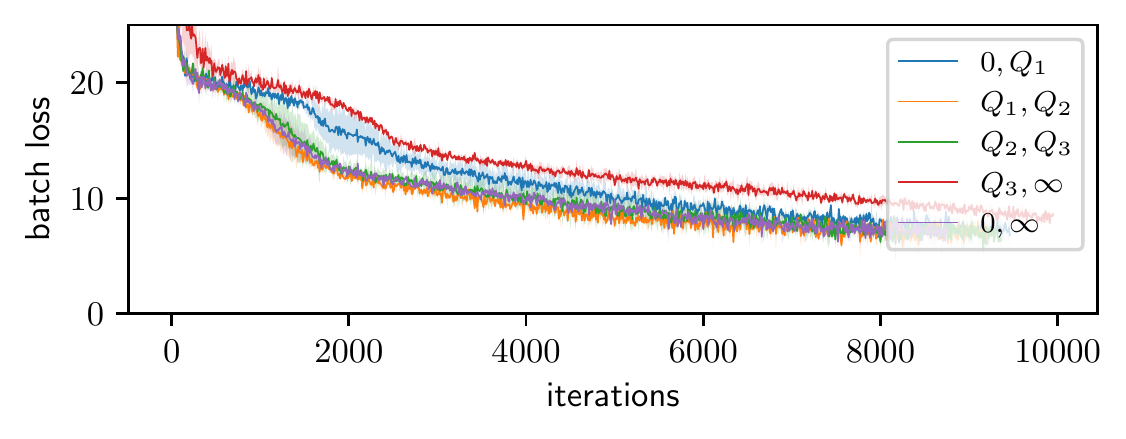}
                \caption{}
        \end{subfigure}
        \caption{
                (a) Histogram for the assembly benchmark consisting of
                1155 demonstrations, equalling 7 hours and 10 minutes of demonstrated strategies in the simulator.
                The dashed lines $Q_1$ to $Q_3$ separate the
                datasets into equally populated quarters.
                (b) Learning curves of the different models,
                trained with $k=4$, $N=25$, learning rate $5 \cdot 10^{-4}$, and
                a minibatch size of $128$.
        }
        \label{fig:dataset_and_learning}
\end{figure}

Training data usually determines the quality of data-driven models.
We thus investigated how the quality of demonstrations influences the learning of assembly strategies.
All demonstrations were recorded at \SI{100}{Hz} from random starting poses until finishing the assembly in simulation.
Since they were all successful, we use their duration as a measure of quality,
implying that the shorter an assembly demonstration took, the higher
the quality of the recorded strategies.
We separated the training set with quartiles $Q_1$ - $Q_3$ as depicted in \fig{fig:dataset_and_learning}(a)
and used five datasets to train on demonstrations with different duration:
\begin{itemize}
        \item $0 \rightarrow Q_1$ : up to \SI{14.93}{s}
        \item $Q_1 \rightarrow Q_2$ : between \SI{14.93}{s} and \SI{18.74}{s}
        \item $Q_2 \rightarrow Q_3$: between \SI{18.74}{s} and \SI{24.32}{s}
        \item $Q_3 \rightarrow \infty$ : longer than \SI{24.32}{s}
        \item $0 \rightarrow \infty$ : all demonstrations
\end{itemize}
\fig{fig:dataset_and_learning}(b) shows the batch loss during training for the different models.
They perform similarly in predicting the labels of the evaluation set with
a slight degradation of the model trained only on the last quarter.

\begin{figure}
        \centering
        \begin{subfigure}[b]{0.40\textwidth}
                \includegraphics[width=1.0\textwidth]{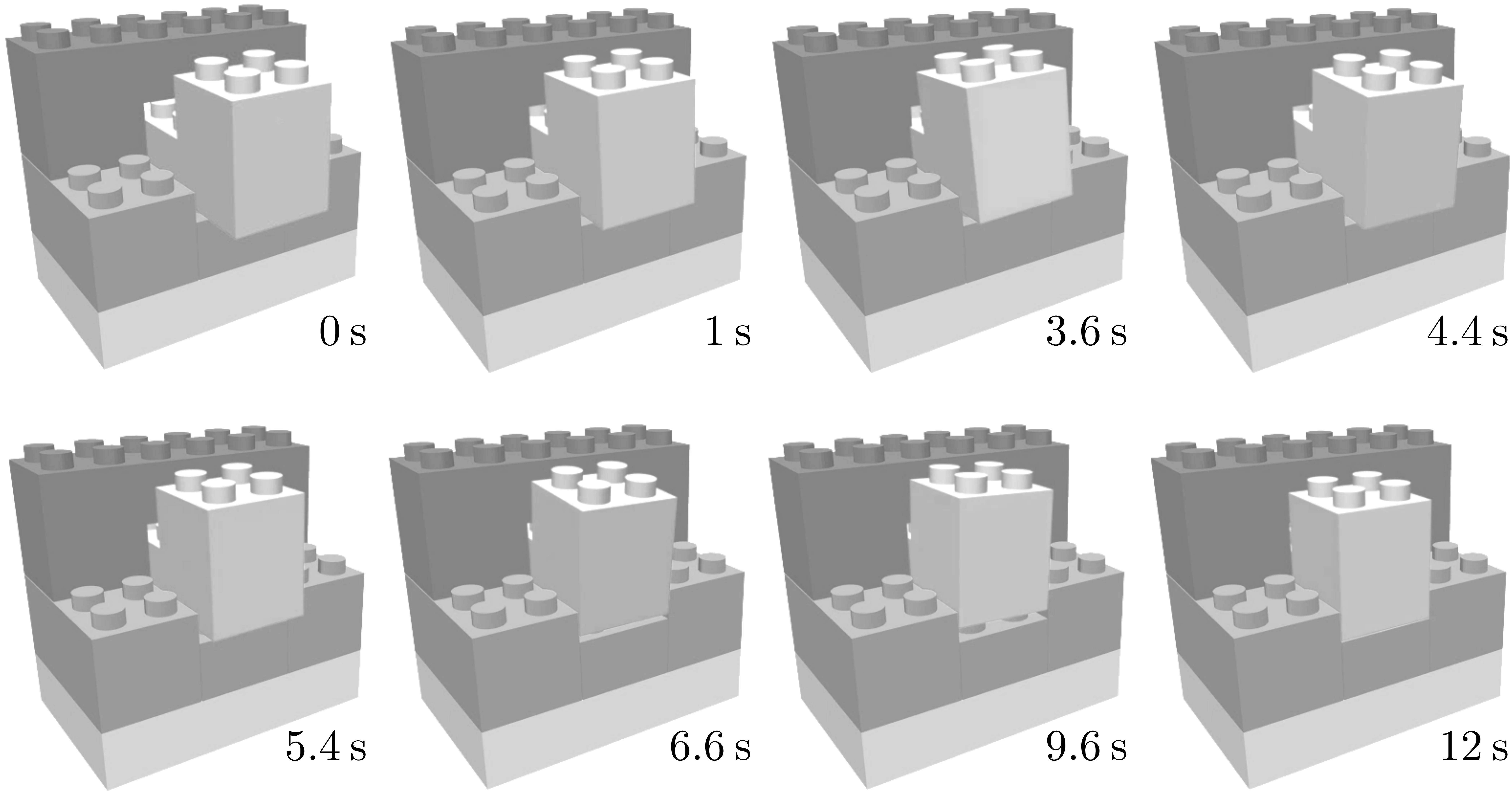}
                \caption{}
        \end{subfigure}
        \begin{subfigure}[b]{0.49\textwidth}
                \includegraphics[width=1.0\textwidth]{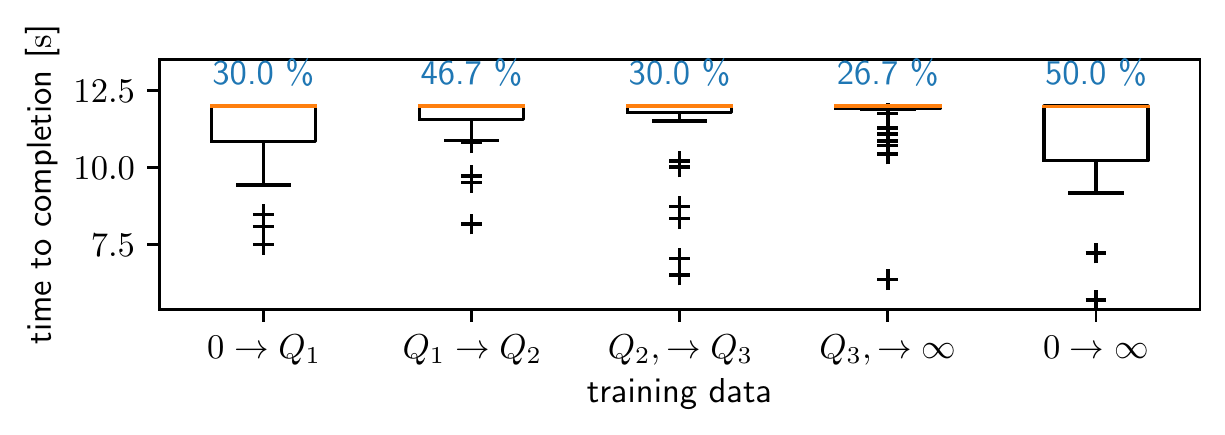}
                \caption{}
        \end{subfigure}
        \caption{
                (a) A successful run in simulation at $20$\% realtime.
                (b) Assembly performance in simulation with the success rate including outliers.
                We tested 30 trials for each model.
                Successful executions reached a goal distance of \SI{0.5}{mm} in {12}{s}.
        }
        \label{fig:simulation_and_success_rate}
\end{figure}

Continuing the experiment, we used the simulator to evaluate the different models on the assembly task.
The models' predicted force-torque strategies steered the assembly
object similar to as was done during demonstration with the joystick.
\fig{fig:simulation_and_success_rate} shows a successful execution.
The assembly has mainly two challenging regions:
The first is the insertion at \SI{3.6}{s} where parts jam and tilt frequently, and the
second is at \SI{6.6}{s} where pressing in requires dexterity due to low-clearances.
We measured success with the Euclidean distance to the goal pose $\Vert \bm{x} \Vert$
and cut off at \SI{12}{s} simulated time (\SI{60}{s} realtime), after which the execution was considered a failure.
\fig{fig:simulation_and_success_rate}(b) shows the results.
The model trained on the complete dataset performs best on the benchmark.
The model trained on the second quarter of the dataset has a
similar performance, which is interesting due to using only $25$\% training data.
This could be due to hitting a sweet spot between data quality and amount of recorded time.
The model trained on the last quartile shows the lowest success rate, corresponding to the degraded learning curve from \fig{fig:dataset_and_learning}.

In conclusion, the quality of demonstrations in the simulator has an impact on
the model to learn successful force-torque strategies and
more effective (shorter) demonstrations can partially replace bigger datasets.

\subsection{Real hardware}
The goal of this experiment was to test if the strategies are portable to real-world systems through our controller.
We prepared this transfer in our concept by allowing partly unrealistic mesh interpenetration in the simulator during
demonstrations and increased friction and stiction between the assembly components.
\begin{figure}
        \centering
        \includegraphics[width=0.47\textwidth]{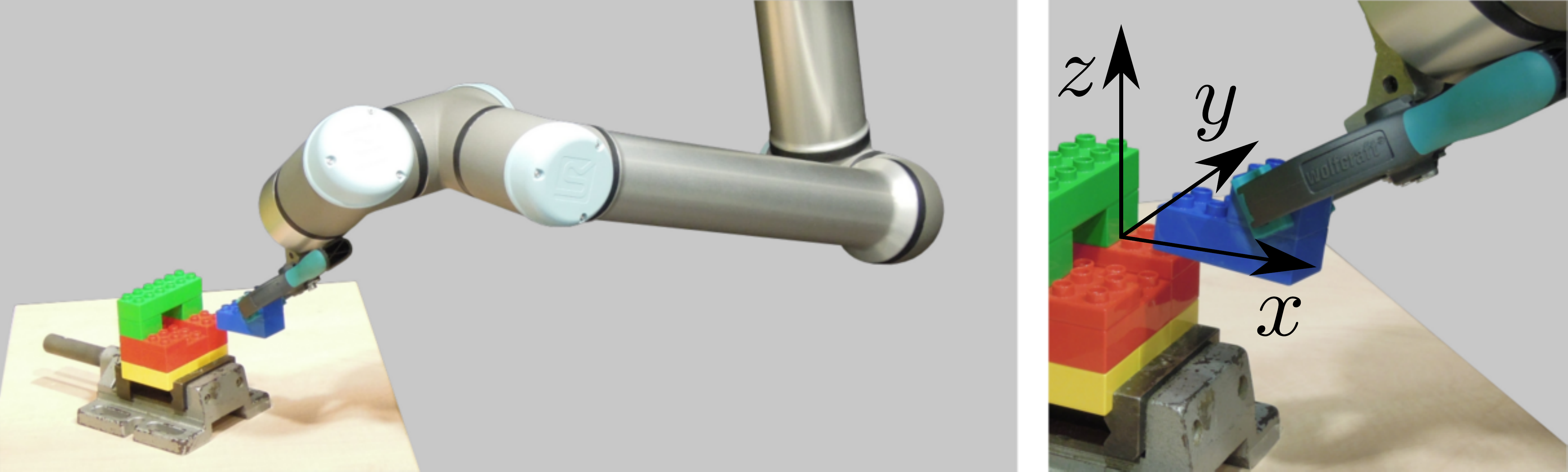}
        \caption{
        Setup for the strategy transfer with the UR10e robot. \textit{Left}: The assembly is
        clamped in a vice and glued to a surface with adhesive tape.
        \textit{Right}: The blue object is glued to the jaws of a non-actuated gripper, whose
        elasticity adds uncertainty to the system.
        }
        \label{fig:assembly_setup}
\end{figure}

\begin{figure}
        \centering
        \includegraphics[width=0.49\textwidth]{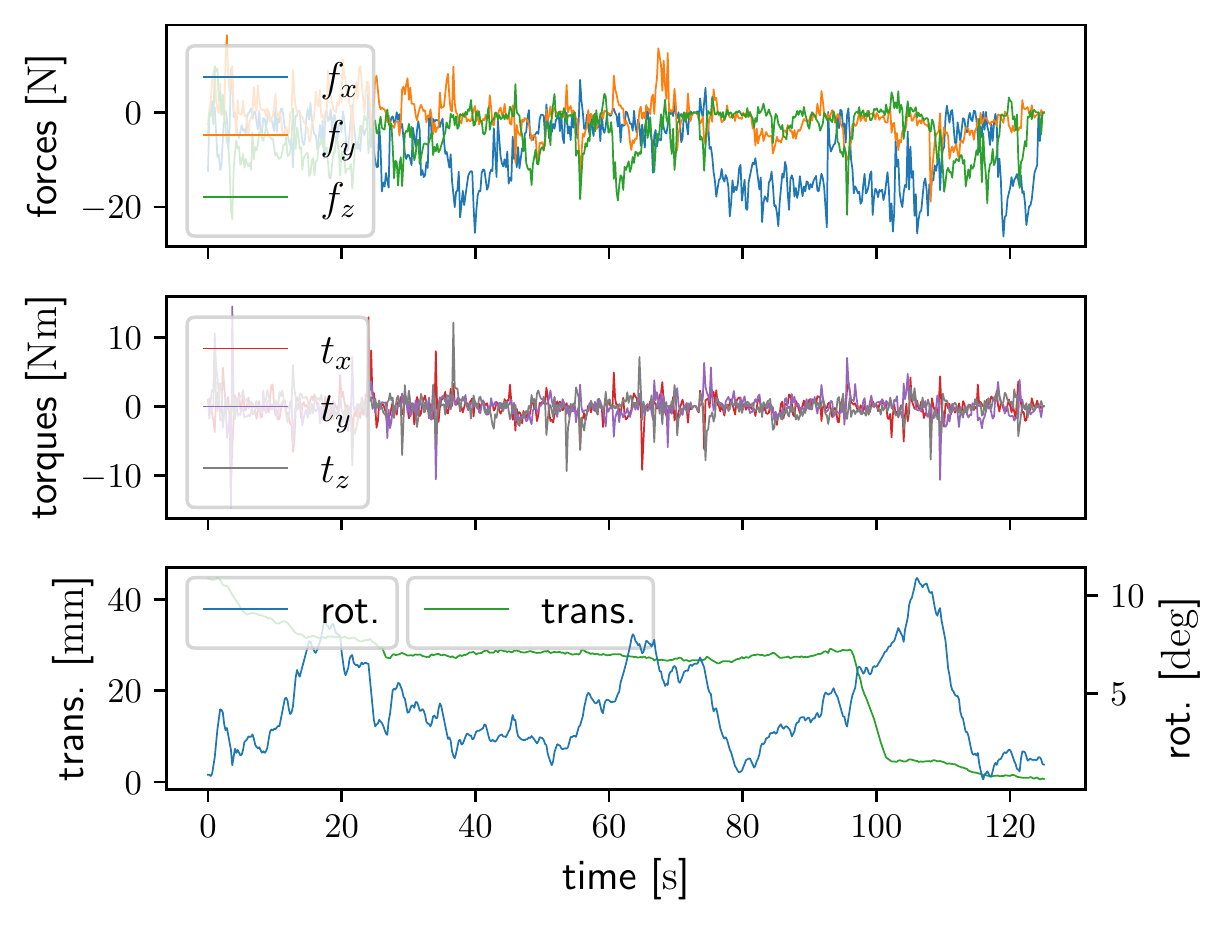}
        \caption{
        Commanded force-torque strategies for a successful assembly on the real setup.
        }
        \label{fig:recorded_assembly_strategies}
\end{figure}

\fig{fig:assembly_setup} shows the setup.
We used a passive gripper for this experiment to attach the blue assembly object to the robot's end-effector.
We calibrated the setup by teleoperating the robot in force
control to the goal pose of the assembly and recorded the final configuration.
During execution, the model generated force-torque reference commands at \SI{5}{Hz}
that were executed with the force controller on the UR10e robot at \SI{500}{Hz}.
\fig{fig:recorded_assembly_strategies} shows the strategies and assembly progress for one such execution.
During the first insertion, the parts quickly jam and require strategies to advance.
Note how the model makes several forward push attempts with the swings of the $f_x$ component.
Since this is not immediately successful, the model tries different re-orientations while searching for part clearance. 
After approximately \SI{95}{s}, the strategies are successful and finally plug both parts together.
The execution is shown in full length in the accompanying video.
\section{Discussion}
\label{sec:discussion}

During the executions on the real setup, configurations that significantly differed
from the dataset decreased the model's performance.
Real-world assemblies might have several local optima along the goal where
parts can be put in unseen and wrong configurations, leading to possible deadlocks.
If these spots are known or can be anticipated, they should be included
randomly as challenges in the simulator to increase the model's robustness.

A further limitation is the speed of execution.
Admittance controlled industrial robots that close force control around
position-controlled joints, are not stable during interaction with stiff environments at human speed.
It might thus be worthwhile to deploy this model together with an impedance-controlled robot.

When programming an assembly with our approach, some experience with the real parts is required
to set suitable ranges for the force-torque signals and to apply meaningful strategies in the simulator.
It is also helpful to touch the real assembly parts to better imagine
how they combine in the simulator.
The strength of our approach is, however, that the predicted strategies can be post-scaled and sanity-checked before
sending them to the controller.
For instance, the force-torque strategies' amplitude could be decreased by $50$\% on the first run.

\section{Conclusions}   %
\label{sec:concludsion}

In this work, we proposed a method to learn human-inspired assembly strategies from
simulated teleoperation.
We addressed two challenges that are common in Imitation Learning for force-sensitive assembly tasks:
Handling low clearances with tight-fitting parts that require much trial and error,
and learning these strategies offline without access to the real hardware.
Basing our ideas on data analysis of recorded demonstrations,
we combined an LSTM-based encoder with an MDN to model probabilistic force
strategies that follow patterns over time.
We evaluated this approach experimentally in simulation and on real hardware
with an exemplary benchmark.
The results showed that human-inspired strategies that are learned from simulated
data with our method are concise enough to be successfully transferred to a real
robot.

In future research, we plan to reduce the sample intensity that is currently required in the simulator.
A promising method could be including Offline RL~\cite{Kostrikov2021} on the
demonstration data to exceed human performance without further interaction with the simulator.

\renewcommand*{\bibfont}{\small}
\printbibliography

\end{document}